\documentclass[a4paper,conference]{IEEEtran}

\usepackage{xcolor}

\usepackage{times}
\usepackage{epsfig}
\usepackage{graphicx}
\usepackage{amsmath}
\usepackage{amssymb}
\usepackage{url}
\usepackage{multirow}
\usepackage{booktabs}
\usepackage{float}
\usepackage{caption}
\usepackage{subcaption}
\usepackage{hyperref}

\def\etal{\textit{et.al.}}

\usepackage{xurl}
\usepackage{fancyhdr}

\begin{document}

\onecolumn This paper is a preprint (Accepted in ICPR 2022).

\hfill \break

\textcopyright  2022 IEEE. Personal use of this material is permitted. Permission from
IEEE must be obtained for all other uses, in any current or future media,
including reprinting/republishing this material for advertising or promotional
purposes, creating new collective works, for resale or redistribution to servers
or lists, or reuse of any copyrighted component of this work in other works.
\twocolumn

\title{Rethinking Task-Incremental Learning Baselines}




\author{
\IEEEauthorblockN{Md Sazzad Hossain\IEEEauthorrefmark{1},
Pritom Saha\IEEEauthorrefmark{1},
Townim Faisal Chowdhury\IEEEauthorrefmark{2},
Shafin Rahman\IEEEauthorrefmark{2}, \\
Fuad Rahman\IEEEauthorrefmark{1} and
Nabeel Mohammed\IEEEauthorrefmark{2} 
}

\IEEEauthorblockA{
\IEEEauthorrefmark{1}Apurba Technologies Limited, Dhaka, Bangladesh. \\
\IEEEauthorrefmark{2}Apurba-NSU R\&D Lab, Department of Electrical and Computer Engineering,\\
North South University, Dhaka, Bangladesh. \\
Email: \{sazzad\_hossain, pritom\_saha, fuad\}@apurbatech.com \\
\{townim.faisal, shafin.rahman, nabeel.mohammed\}@northsouth.edu 
}

}



\maketitle

\begin{abstract}
    It is common to have continuous streams of new data that need to be introduced in the system in real-world applications. The model needs to learn newly added capabilities (future tasks) while retaining the old knowledge (past tasks). Incremental learning has recently become increasingly appealing for this problem. Task-incremental learning is a kind of incremental learning where task identity of newly included task (a set of classes) remains known during inference. A common goal of task-incremental methods is to design a network that can operate on minimal size, maintaining decent performance. To manage the stability-plasticity dilemma, different methods utilize replay memory of past tasks, specialized hardware, regularization monitoring etc. However, these methods are still less memory efficient in terms of architecture growth or input data costs. In this study, we present a simple yet effective adjustment network (SAN) for task incremental learning that achieves near state-of-the-art performance while using minimal architectural size without using memory instances compared to previous state-of-the-art approaches. We investigate this approach on both 3D point cloud object (ModelNet40) and 2D image (CIFAR10, CIFAR100, MiniImageNet, MNIST, PermutedMNIST, notMNIST, SVHN, and FashionMNIST) recognition tasks and establish a strong baseline result for a fair comparison with existing methods. On both 2D and 3D domains, we also observe that SAN is primarily unaffected by different task orders in a task-incremental setting.
\end{abstract}


%
\IEEEpeerreviewmaketitle

\section{Introduction} \label{sec:introduction}


Task-incremental learning is a brain-inspired process to incrementally learn a set of tasks without forgetting previously known knowledge. Each task includes several classes, and task identities remain known during the testing stage. For example, a vehicle detection system may need to adapt itself to detect pedestrians, trees, buildings, etc., as a new task in addition to vehicles. Instead of training from scratch, task-incremental learning augments the new knowledge without compromising the past experience. It is an active research area in continual or lifelong learning. This paper proposes a simple yet effective method for task-incremental learning that can serve as a strong baseline while developing any technique.

\begin{figure}[!t]
\centering
\includegraphics[width=\linewidth,trim={0cm 0.0cm 0cm 0cm},clip]{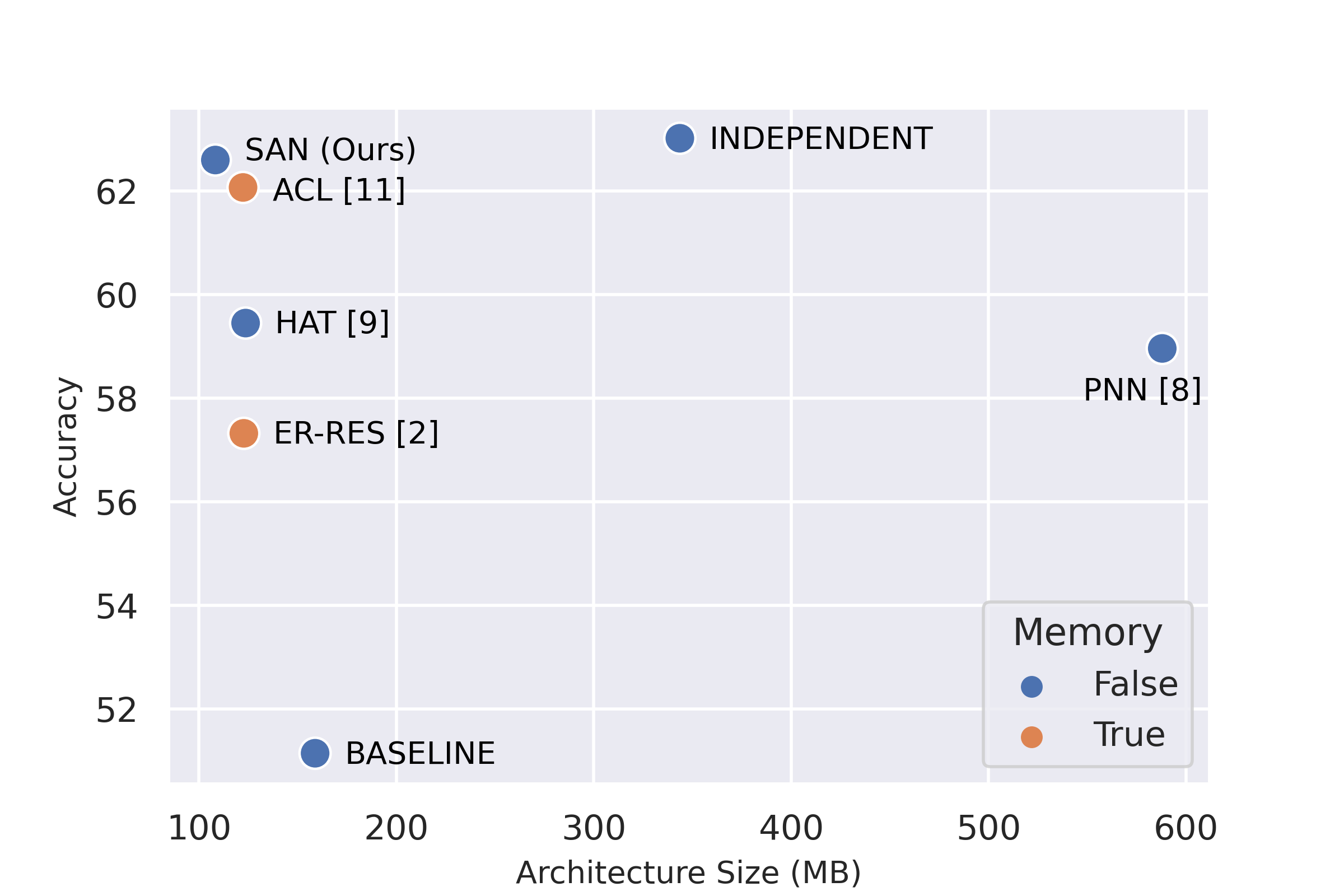}
\caption{Accuracy vs. model size comparison among existing and SAN (Ours) method on the 20-split MiniImagenet dataset. SAN (Ours) attains a perfect accuracy-size trade-off even without reply memory examples.
}
\label{fig:comparison}
\end{figure}

The main challenge of task-incremental learning is to maintain a reasonable performance using a small model size (i.e., less trainable parameters). In doing so, recent approaches of applies different strategies for task-incremental learning.
\textit{\textbf{(1)}} Replay memory: Methods like \cite{buzzega2021rethinking, chaudhry2019tiny, guo2019improved, weston2014memory, pritzel2017neural, santoro2016one} store multiple samples from previous tasks while learning a new task to control catastrophic forgetting. It increases the architecture size because the memory grows exponentially with the number of tasks upraises. 
\textit{\textbf{(2)}} Specialized architectures: Some methods increase the capacity of the network by employing specialized architectures like autoencoders. For example, \cite{aljundi2017expertgate} uses autoencoders during testing to examine representations of previous tasks and select experts for specific tasks. It requires a significant amount of time to train, and it also increases the inference time. 
\textit{\textbf{(3)}} Regularization based methods \cite{chaudhry2018riemannian, ebrahimi2019uncertainty, serra2018overcoming} adapt the new incremental data by regularizing changes to parameters for upcoming tasks. Although this method does not require additional replay memory, it requires continuous parameter observation and the learning rate adjustment that is computationally expensive and time-consuming.
This paper addresses the accuracy and model size trade-off using a Simple Adjustment Network (SAN), bypassing the complex strategies used in earlier work. Fig. \ref{fig:comparison} provides an accuracy vs. model size comparison for task-incremental learning methods. Our proposed method successfully makes a perfect balance of accuracy vs. model size vs. memory requirement.

This work introduces a simple adjustment network (SAN) approach that utilizes knowledge from the previous tasks. It has three main components: the backbone, simple adjustment network, and classifier. For the first task, we train the entire network using the training data of the first task. The following incremental tasks use the trained backbone and classifier on the first task and only learn the adjustment layers for all upcoming tasks. Freezing the rest of the network makes the training process simple, efficient, and compact. Moreover, SAN does not require any episodic memory or examples to revise previous data to avoid forgetting. As task-incremental learning allows using task identity during inference time, the model selects a specific adjustment network trained for that task. Despite its simplicity, this approach can compete with the most complex works of task incremental learning. Hence, In terms of performance, architecture size, runtime, catastrophic forgetting, SAN can serve as the baseline for task-incremental learning. We validate our proposal using both 2D image and 3D point cloud datasets. Our experiments show consistent results beating many state-of-the-art methods (PNN \cite{rusu2016progressive}, HAT \cite{serra2018overcoming}, ER-RES \cite{chaudhry2019tiny}, A-GEM \cite{chaudhry2018efficient}, ACL \cite{ACL_ECCV_2020}, UCB \cite{ebrahimi2019uncertainty}) of task-incremental learning. In summary, the contributions are:
\begin{itemize} 
    \item We propose a novel and simple yet effective baseline approach for task-incremental learning that promises less model size maintaining superior performance than state-of-the-art methods.
    \item Our proposed method can work without replay memory or exemplar data of previous tasks.
    \item To the best of our knowledge, we are the first to test task-incremental learning on 3D point cloud objects (ModelNet40 \cite{modelnet2015}). We have also performed extensive experiments on 2D image data (CIFAR10, CIFAR100 \cite{cifar100}, MiniImageNet \cite{miniimagenet}, MNIST \cite{lecun-mnisthandwrittendigit-2010}, PermutedMNIST \cite{lecun1998gradient}, SVHN \cite{netzer2011reading} and FashionMNIST \cite{xiao2017fashion}).
\end{itemize}


\section{Related Works}
\label{sec:related}



\noindent\textbf{Task-incremental learning:} Continual learning addresses learning progressively from an endless sequence of tasks while avoiding catastrophic forgetting. Many methods \cite{10.1007/978-3-319-46493-0_37,8100070,kemker2018fearnet,v.2018variational}  have been proposed to overcome the challenges of continual learning. We solely consider task-incremental learning as continual learning because of its simplicity. There are three bodies of work for task-incremental learning:  memory-, regularization-, and architecture-based methods. 
\textit{Memory-based methods} \cite{buzzega2021rethinking, chaudhry2019tiny, guo2019improved, weston2014memory, pritzel2017neural, santoro2016one, liu2020mnemonics} use rehearsal or pseudo rehearsal strategy to mitigate catastrophic forgetting. To maintain knowledge of previous tasks, the rehearsal strategy \cite{lopez2017gradient,buzzega2021rethinking, chaudhry2019tiny} employs a set of old input exemplars stored in memory and replayed throughout incremental task training. In contrast, pseudo rehearsal strategy \cite{xiang2019incremental, lee2019generative} uses a generative model to synthesize the old inputs. However, the disadvantages of memory-based approaches are that they are primarily reliant on knowledge from previous tasks and are less memory efficient, reduce model plasticity, and cannot capture the whole dataset distribution. In our work, we do not store any samples as reply memory.
\textit{Architecture-based methods} \cite{rusu2016progressive, wang2017growing, aljundi2017expertgate, serra2018overcoming, chaudhry2018efficient, lee2019overcoming, ACL_ECCV_2020} employ ways for sequentially acquiring new knowledge from new tasks by expanding the size of the deep learning model. The advantage of this method is not to use any exemplar memory. Rusu \etal{} \cite{rusu2016progressive} proposed Progressive Neural Network (PNN) that grows the model statistically while lateral connections are learned to mitigate the effect of catastrophic forgetting. Yoon \etal{} \cite{yoon2018lifelong} introduced dynamically expandable networks (DEN), which may increase the scale of the architecture based on the interdependency of knowledge across tasks. However, the key difficulty of these methods is to optimize the impact of model growth, maintaining decent performance. 
\textit{Regularization-based methods} \cite{mallya2018piggyback, tang2020graph, riemer2018learning, kirkpatrick2017overcoming, zenke2017continual, chaudhry2018riemannian, fernando2017pathnet, ebrahimi2019uncertainty, 10.1007/978-3-319-46493-0_37} reduce memory requirements while introducing a regularization term in loss function or controlling the parameters of the model. Li \etal{} \cite{ 10.1007/978-3-319-46493-0_37} proposed to use a knowledge distillation loss to retain old task's knowledge while learning new task's knowledge. However, this method suffers when data shift occurs. In EWC \cite{kirkpatrick2017overcoming}, Kirkpatrick \etal{} introduced a new penalty term in loss function to calculate essential parameters using the diagonal of the Fisher information matrix. Aljudni \etal{} \cite{aljundi2018memory} suggested an unsupervised method for estimating the significance of weights in the model. The main challenge of the aforementioned methods is to avoid the stability-plasticity dilemma of the model if the sequence of tasks is long.\\
\noindent\textbf{Task incremental learning baselines:} Different approaches choose different baseline techniques to evaluate their findings. Regularization-based methods such as UCB \cite{ebrahimi2019uncertainty} use finetuning, joint training, and feature extraction methods as baselines. Finetuning method trains a new set of data with the initialization of previously learned parameters. Joint training combines all datasets and trains them as a single model. Feature extraction freezes all layers except the last layer after training to the first task. HAT \cite{serra2018overcoming} compares their results with SGD \cite{goodfellow2013empirical} and a modified version of SGD. On the other hand, memory- and architecture-based methods like Growing a Brain \cite{wang2017growing}, PNN \cite{rusu2016progressive}, ER-RES \cite{chaudhry2019tiny}, MEGA \cite{guo2019improved} use finetuning as their primary baseline. Also, these methods consider EWC \cite{kirkpatrick2017overcoming}, LFW \cite{10.1007/978-3-319-46493-0_37}, and VAN as other baseline methods. A-GEM considered GEM \cite{lopez2017gradient} as baseline since A-GEM is build upon \cite{lopez2017gradient}. Although finetune is the most prominent transfer learning method, this method suffers from catastrophic forgetting and is not suitable for incremental learning. The feature extraction method does not learn anything after the first task. ECW and LFW follow an incremental manner but cannot overcome the forgetting issue. This paper proposes a simple baseline method for incremental tasks using previous knowledge with bare minimum architecture growth and no memory dependency.\\
\noindent\textbf{Incremental learning on 3D point cloud:} This is a relatively unexplored area in the context of incremental learning. Recent efforts, \cite{liu2021l3doc, chowdhury2021learning} applied incremental learning on 3D point cloud data. Liu \etal{} \cite{liu2021l3doc} used memory attention mechanism during incremental learning of 3D point cloud data. Chowdhury \etal{} \cite{chowdhury2021learning} proposed to use semantic information alongside feature representations of 3D point cloud data to minimize catastrophic forgetting in incremental learning. However, none of the methods addressed task-incremental learning.

\begin{figure}[!t]
\centering
\includegraphics[width=.5\textwidth,trim={0cm 0cm 0cm 0cm},clip]{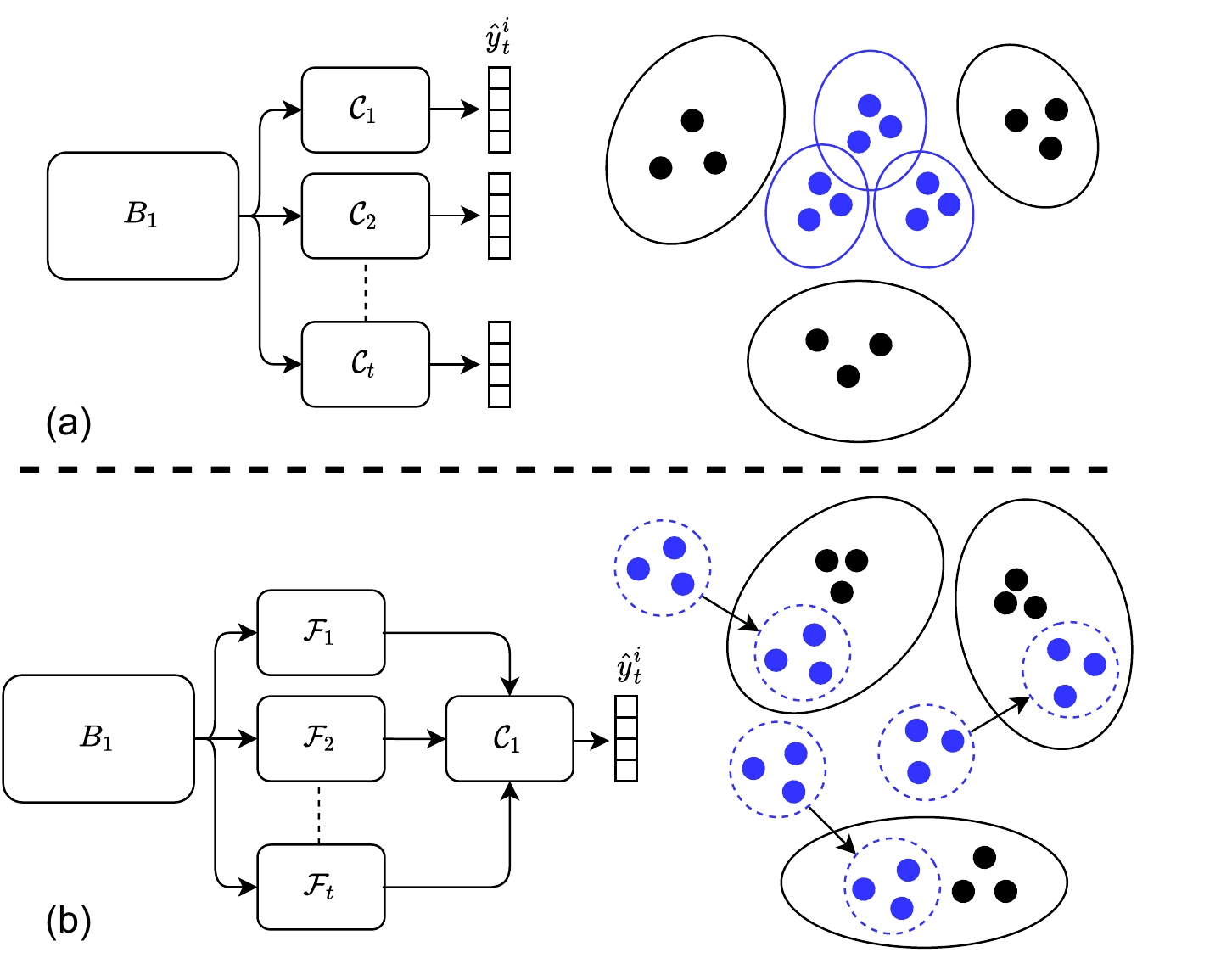}
\caption{\textbf{(a)} Baseline model. New classifier $\mathcal{C}_t$ (FC layers) is added every time when new tasks arrive. It creates new decision boundaries (in \textcolor{blue}{blue}) in latent space. Newer decision boundaries may not be as good as the first one (in black) because the backbone $B_1$ remains frozen during incremental steps. \textbf{(b)} Our proposed SAN model. New adjustment network $\mathcal{F}_t$ (conv. layers) is added at each incremental step keeping the classifier $\mathcal{C}_1$ fixed after the first incremental step. $\mathcal{F}_t$ supports the backbone, $B_1$ to produce better features and align new features (arrow) with previously built classifier, $\mathcal{C}_1$.
}
\label{fig:motivation}
\end{figure}

\section{Task-incremental Learning}
\label{sec:methodology}

\noindent \textbf{Problem formulation:}
Assume, a set of tasks, $\mathcal{T} = \{\tau_1, \tau_2, ..., \tau_{|\mathcal{T}|}\}$, where $|\mathcal{T}|\geq1$ has been introduced to a model $\mathcal{M}$ sequentially. Each task $\tau_t$ consists of input set $\mathcal{X}_{t}$ with label set $\mathcal{Y}_{t}$, where $\mathcal{X}_{t} = \{x^i_t\}_{i=1}^{n_t}$, $x^i_t$ 
is the $i$-th instance with class label $y^{i}_{t}\in\mathcal{Y}_t$ and $n_t$ is the total number of instances included in task $\tau_t$. There is no overlap among the classes of different tasks, i.e., $\forall i,j, \mathcal{Y}_{i} \cap  \mathcal{Y}_{j}=\varnothing$. 
We design an object classification model $\mathcal{M}$ using the source data i.e., input set $\mathcal{X}_1$ with corresponding label set $\mathcal{Y}_1$ from task $\tau_1$ and termed this model as \textit{base model}. Our aim is to incrementally build this model $\mathcal{M}$ and train with target data i.e., non-overlapped inputs from the task, $\{\tau_{t}\}_{t=2}^{|\mathcal{T}|}\in\mathcal{T}$, while retaining all knowledge from previous tasks. An important consideration for task-incremental learning is that the task identity, $\tau_t$ remains known during testing the classes related to the task $t$.
\subsection{Solution overview}\label{sec:solution-overview}
We first discuss a common/naive baseline method usually used in literature. After that, by mentioning the issues related to this baseline, we outline our proposed approach, which could serve as a more robust baseline.

\noindent\textbf{Common baseline:} The model consists of a baseline network, $B_1$ (convolutional layers) and task specific classifiers, $\mathcal{C}_t$ (fully connected layers). A softmax activation is placed the end of $\mathcal{C}_t$. At every $t$th incremental step, a new classifier $\mathcal{C}_t$ is added parallel to $\mathcal{C}_{t-1}$ as a new classification head for the prediction score of classes related to task $t$. The first incremental training begins from scratch with $B_1$ and $\mathcal{C}_1$ using base task (Task 1) data, $(\mathcal{X}_{1},\mathcal{Y}_{1})$. Only at this step, the full network remains trainable. Later, at $t(>1)$th incremental step, backbone, $B_1$ remains frozen but classifier $\mathcal{C}_t$ receives training using target data, $(\mathcal{X}_{t},\mathcal{Y}_{t})$. As the task identify is known during testing, the network can choose the corresponding classifier $\mathcal{C}_t$ in addition to the common backbone, $B_1$ for prediction. The overall calculation to obtain the prediction scores $(\hat{y}_t^i)$ for an instance $x_t^i$ belonging to $t$th task can be summarized as:
\begin{equation}
    \hat{y}_t^i = \mathcal{C}_t\big( B_1(x_t^i;\theta_b);c_t \big)
\end{equation}
where, $\theta_b$ and $c_t$ are the parameters of $B_1$ and $\mathcal{C}_t$, respectively. Fig \ref{fig:motivation} (a) illustrates this model and its feature space. At every incremental step newer class boundaries are created based on number of class added in $\mathcal{C}_t$. 


\noindent\textbf{Key challenges:} Task 1 training at $t=1$ generates well-separated and regularized class boundaries because full network i.e., both $B_1$ and $\mathcal{C}_1$ are jointly trained with Task 1 data. However, for the tasks $t > 1$, $B_1$ is not updated with the target data. Therefore, only updating $\mathcal{C}_t$, the class boundaries related to task $t > 1$ could not perfectly generalize on training data. It results in a good performance for Task 1 but related less performance for task $t$ where $t>1$. 

\begin{figure}[t]
\centering
\begin{subfigure}[t]{0.47\columnwidth}
    \centering
    \includegraphics[width=\textwidth]{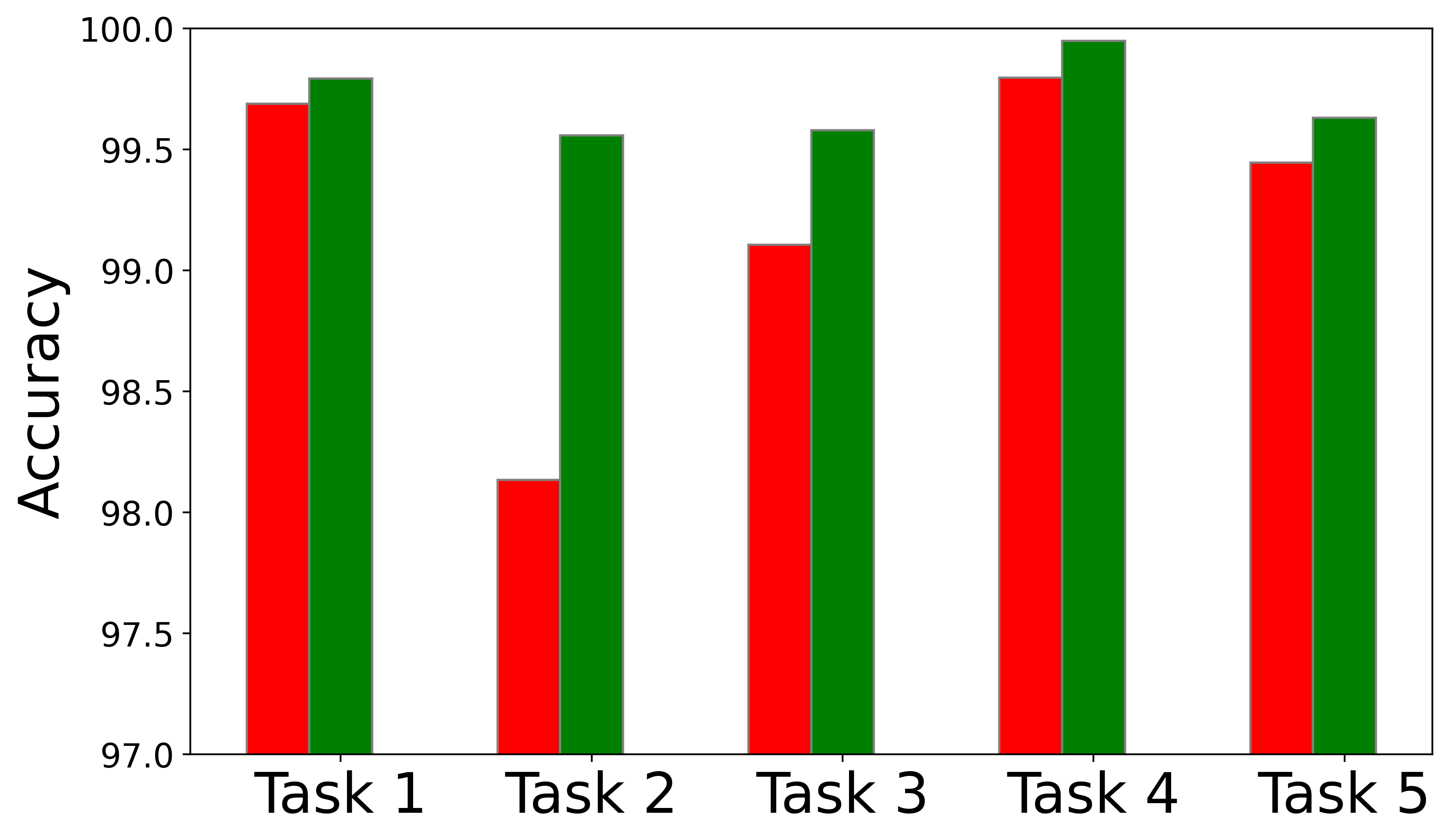}
    \caption{5 Split MNIST}
    \label{PWM_with_DK}
\end{subfigure}
~ 
\begin{subfigure}[t]{0.47\columnwidth}
    \centering
    \includegraphics[width=\textwidth]{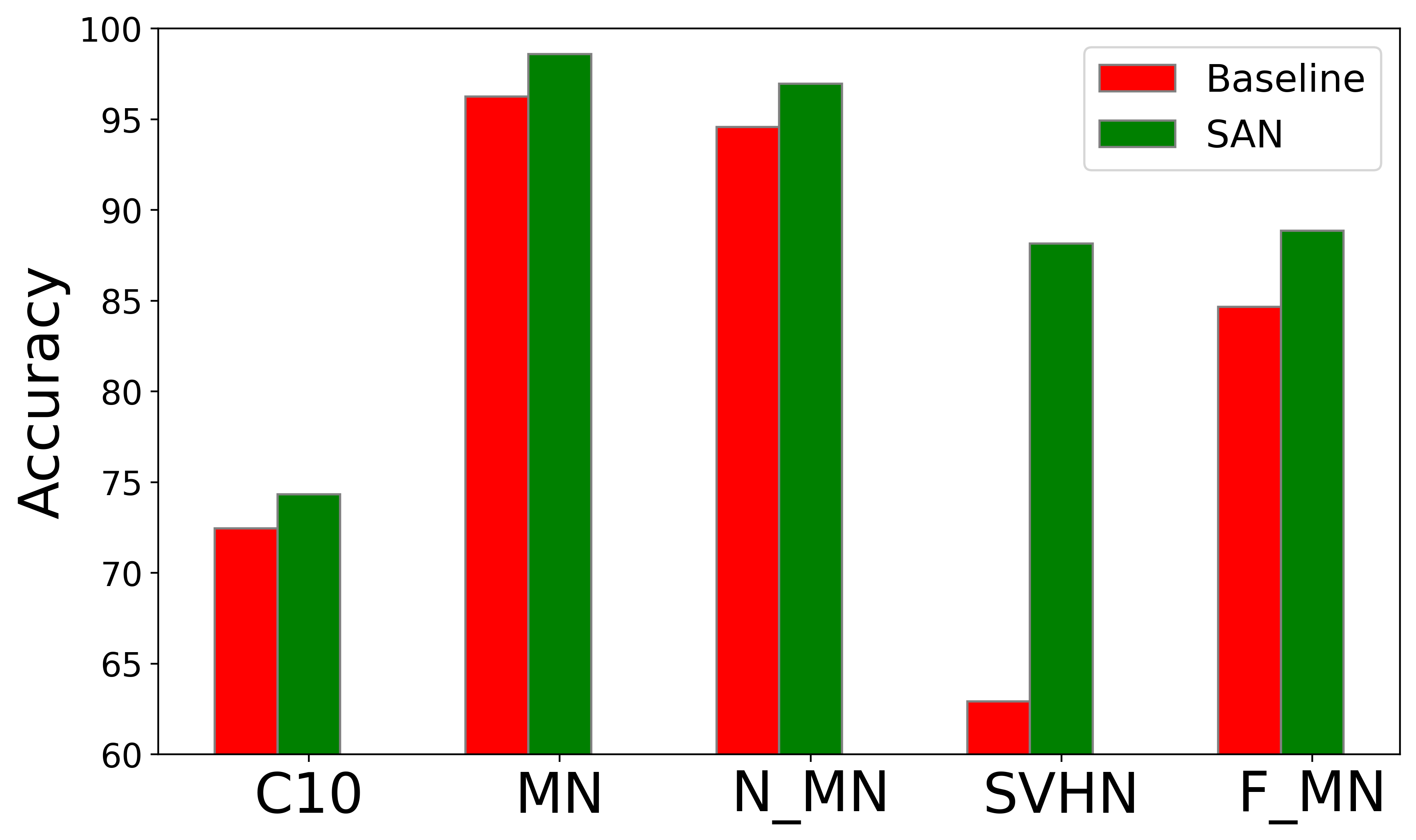} 
    \caption{Sequence of 5 Datasets}
    \label{GameRV_with_DK}
\end{subfigure}
~ 
\begin{subfigure}[t]{0.47\columnwidth}
    \centering
    \includegraphics[width=\textwidth]{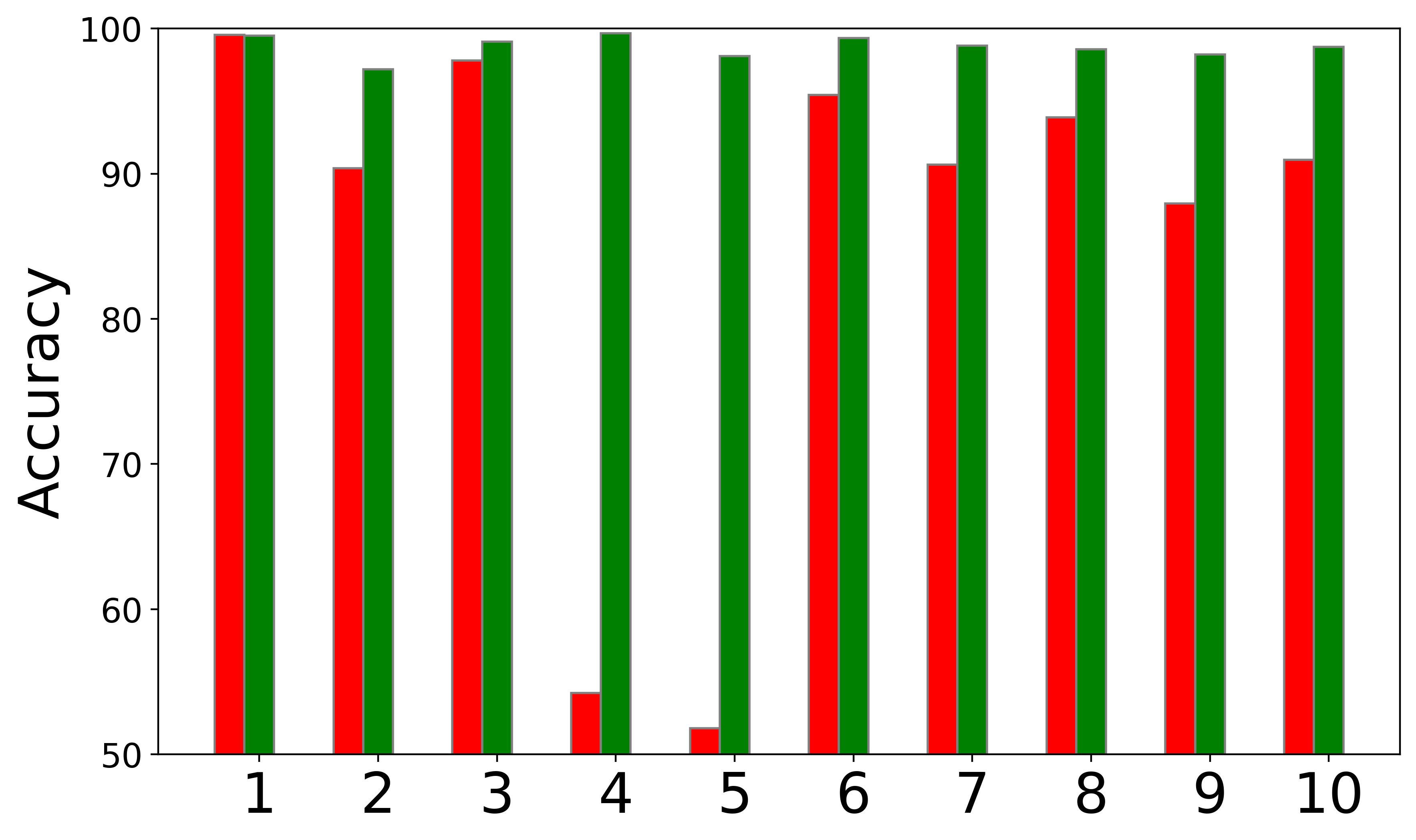}
    \caption{Permuted MNIST}
    \label{PWM_anti_DK}
\end{subfigure}
~ 
\begin{subfigure}[t]{0.47\columnwidth}
    \centering
    \includegraphics[width=\textwidth]{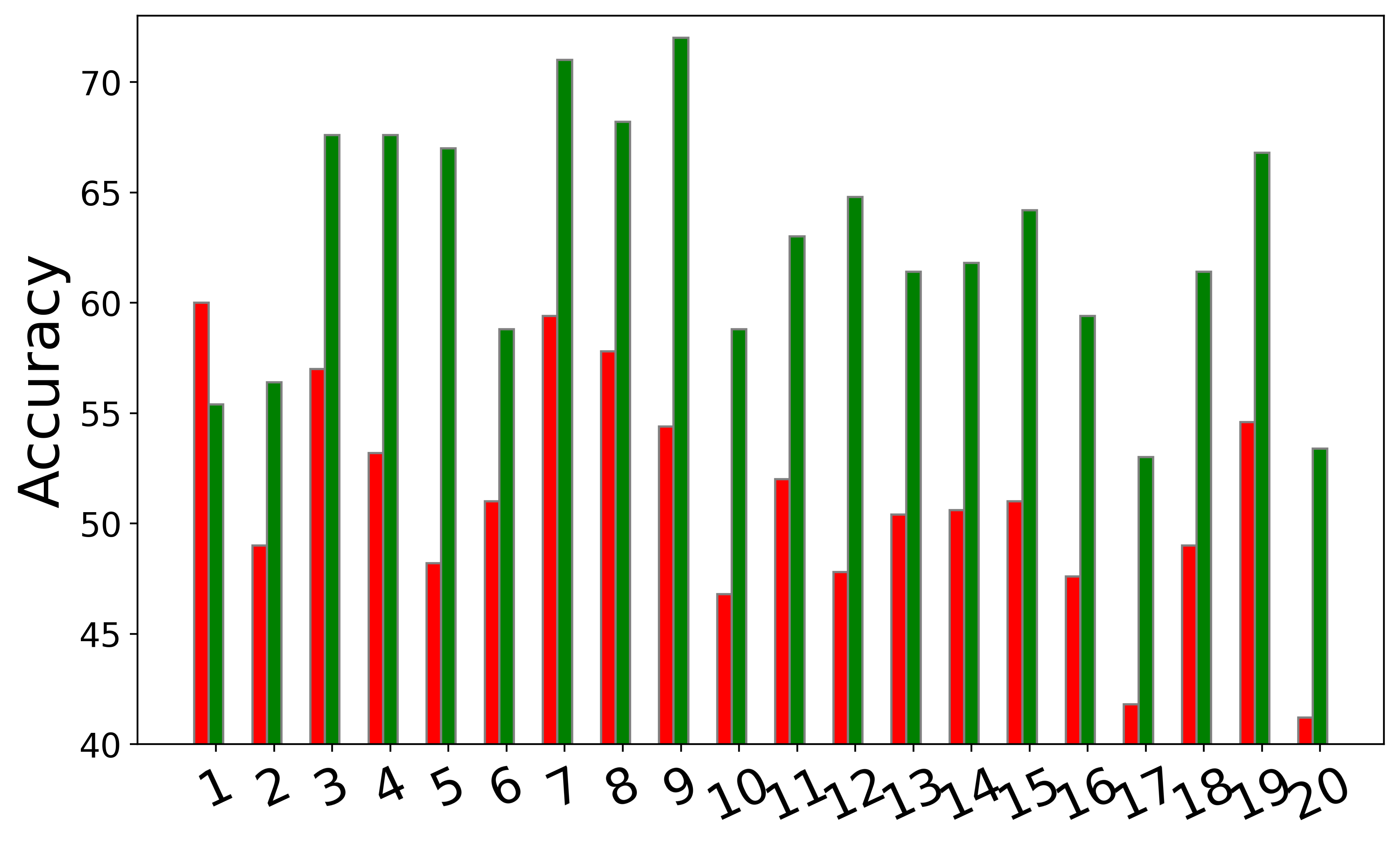}
    \caption{20 Split MiniImagenet}
    \label{PWM_anti_DK}
\end{subfigure}
\caption{Individual task results of common baseline and SAN. One can notice SAN works better than the baseline when more and more tasks are added incrementally.} \label{fig:individual_task}
\end{figure}
\noindent\textbf{Our proposal:} We show our proposal in Fig. \ref{fig:motivation} (b). We propose a simple adjustment network (SAN), $\mathcal{F}_t$ (implemented as a few conv. layer) placed in between the backbone, $B_1$ and classifier, $\mathcal{C}_1$. In this design, $B_1$ and $\mathcal{C}_1$ receive training using source data at Task 1 and remain fixed in other incremental steps. Unlike the common baseline, here, a $\mathcal{F}_t$ is added to the model in each incremental step and receives training with individual tasks' data. During Task 1 training, as it trains the full network, it finds well-separated and regularized class boundaries for Task 1 classes. As no new classifier is added in other incremental steps, target class data only align itself (by training $\mathcal{F}_t$) to the class boundaries formed during Task 1. The overall calculation can be summarized as:
\begin{equation} \label{eq:proposed_san}
    \hat{y}_t^i = \mathcal{C}_1\Big(\mathcal{F}_t\big( B_1(x_t^i;\theta_b); \theta_t \big);c_1 \Big)
\end{equation}
where, $\theta_b$, $\theta_t$ and $c_1$ are the parameters of $B_1$, $\mathcal{F}_t$ and $\mathcal{C}_1$, respectively. This design has several benefits over the common baseline: \textbf{(1)} SAN incrementally adds convolutional layers, $\mathcal{F}_t$ instead of fully connected layers of the common baselines. Training $\mathcal{F}_t$ in each incremental step helps to learn class-specific features, which eventually contributes to better intermediate task performance ($t > 1$) than common baselines (see Fig. \ref{fig:individual_task}). Using the same classifier, $\mathcal{C}_1$ (formed using Task 1 classes) for every incremental step does not create any problem because $\mathcal{C}_1$ is already sufficiently distinctive, and other class instances only need to align the data to cluster under the boundaries of $\mathcal{C}_1$. \textbf{(2)} Being convolutional layers, $\mathcal{F}_t$ has less trainable parameters than fully connected layers, which reduces the model size. \textbf{(3)} This solution can still support a different number of classes in successive incremental steps. If the number of classes to be added in any incremental steps is less than that of Task 1, some part of the classifier $\mathcal{C}_1$ will remain unused (or pruned). In the opposite case, new neurons could be added in $\mathcal{C}_1$ to support exceeded number of classes. Newly added weights have to be trained using new data.

\subsection{Case study 1: With Image data} \label{section:case_study_image_data}

\noindent\textbf{Architecture:} 
For 2D images, we use a convolutional neural network architecture shown in Fig. \ref{fig:architecture}(a). As input, the network takes an image of shape (3, 32, 32) and passes it to the backbone $B_1$. The backbone consists of 3 convolution layers and a maxpool layer. After that, features are passed through a feature encoder, denoted as $\mathcal{F}_t$ containing 4 convolution layers and a maxpool layer. Finally, the flatten feature embedding $\mathbf{f}$ is fed through a fully-connected classifier layer $\mathcal{C}_1$. As an activation function, all layers utilize ReLU. 

\noindent\textbf{Training:} 
A batch size of 64 is used during training with a learning rate of 0.001 for 30 epochs. Adam is used for optimizing the network during the training procedure based on the cross-entropy loss $L_{CE}$ elaborated in Eq. \ref{eq:cross-entropy}. 
\begin{equation} \label{eq:cross-entropy}
    L_{CE} = -\frac{1}{N} \sum_{i=1}^{N}\mathbf{y}^{n}_{i}log(\mathbf{\hat{y}}^{n}_i)
\end{equation}
As task-incremental learning requires task identity $t$ at inference, the model switches the feature encoder $\mathcal{F}_t$ based on $t$ and utilizes $B_1$ and $\mathcal{C}_1$ for prediction as described in Eq. \ref{eq:proposed_san}.

\subsection{Case study 2: With 3D point cloud data}
\begin{figure}[!t]
\centering
\includegraphics[width=.48\textwidth,trim={.25cm 0cm .8cm .3cm},clip]{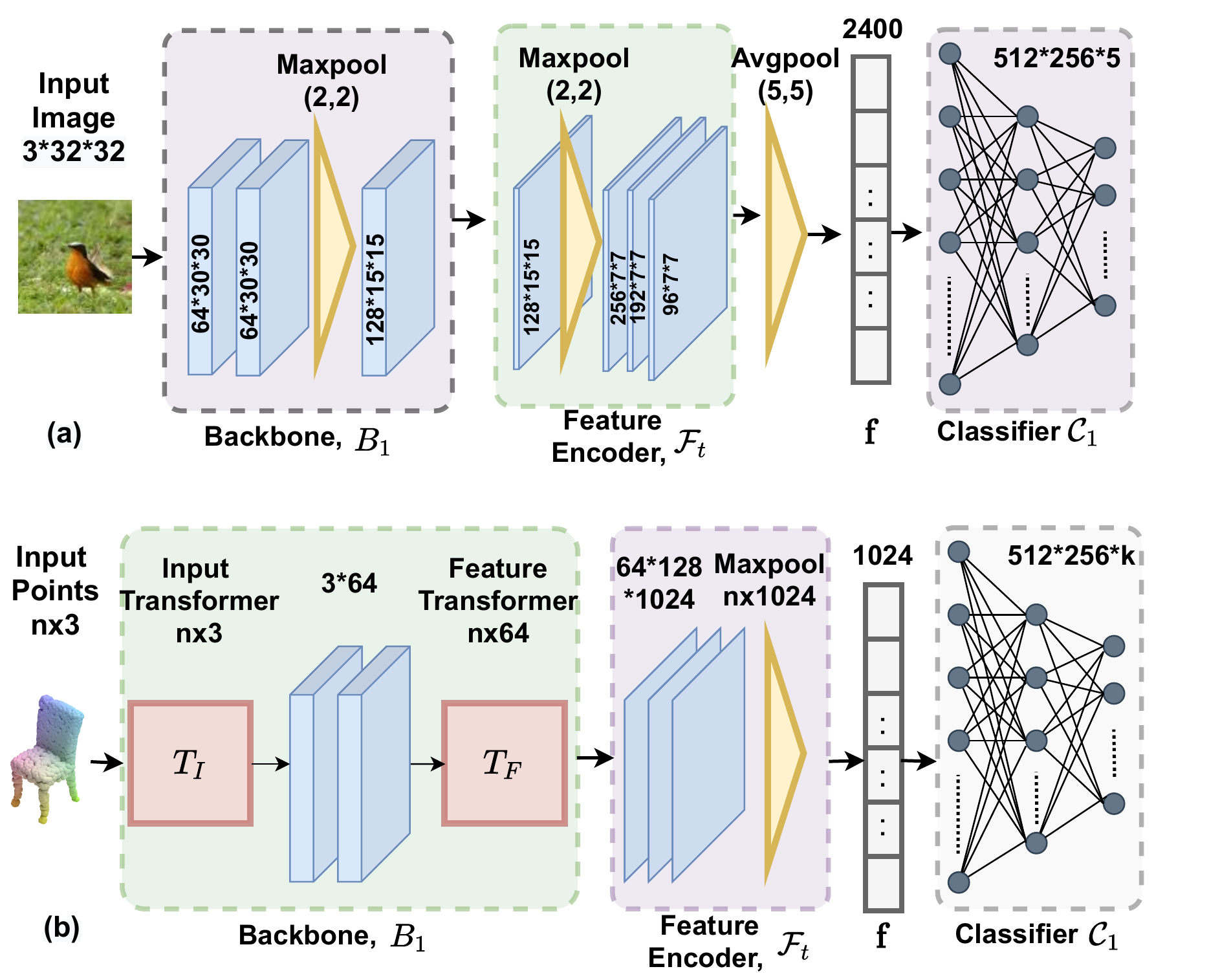}
\caption{Architecture for \textbf{(a)} 2D image and \textbf{(b)} 3D point cloud. After training based on Task 1, Backbone, $B_1$ and classifier, $\mathcal{C}_1$ remain freezed for rest of the training, while adjustment network, $\mathcal{F}_t$ receives training based on the data of Task $t$.}
\label{fig:architecture}
\end{figure}

\noindent\textbf{Architecture:} We utilize PointNet \cite{pointcnn2018} as a 3D point cloud architecture, shown in Fig. \ref{fig:architecture}(b). The model takes $n$ points of a point cloud as input and passes it to the backbone, $B_1$ of the model. The backbone is composed of an input transformer ($T_I$), shared multi-layer perceptron layers (MLP) of size (3, 64), and a feature transformer ($T_F$). Both input and feature transformers have the same architecture and provide pose normalization to any geometric transformations of a point cloud object. The backbone outputs a $64\times64$ feature transformation matrix, $\textit{A}$, which is sent into the feature encoder, $\mathcal{F}_t$. The feature encoder is a shared MLP (64, 128, 1024) network combined with a max-pooling layer that generates the feature embedding, $\textbf{f}$ of the point cloud object. This feature embedding is then passed through two fully connected layers of size 512 and 256, denoted as the classification layer, $\mathcal{C}_1$. Here, all layers are used batch normalization with ReLU.

\noindent\textbf{Training:} Like the training of image data in Sec. \ref{section:case_study_image_data}, the training procedure is same for 3D data except the loss function. The overall loss to train this model is:
\begin{equation}
    L = L_{CE} + \alpha L_{R}
\end{equation}
where, a regularization loss, $L_{R} = \left\| I-AA^T \right\|^2_F$ (with a fixed $\alpha=0.001$) is added to a softmax cross-entropy loss, $L_{CE}$. $L_{R}$ aids in the approximation of the feature transformation matrix, $\textit{A}$ to orthogonality and prevents overfitting of transformers.


\begin{table*}[!t]
\small
\begin{center}
\scalebox{.9}{
\begin{tabular}{@{}lllllll@{}}
\toprule
Experiments                                 & 20 Split CIFAR-100 & 20 Split MiniImagenet & 5-Split MNIST & Permuted MNIST & Sequence of 5 Datasets & 4 split ModelNet40\\ \midrule
Fine-Tune                                      & 34.71 (27.2)       & 28.76 (37.6)          & 65.96 (1.1)   & 44.91 (1.1)    & 27.32 (16.5)           & 78.39 (13.61)\\
HAT \cite{serra2018overcoming}              & 76.96 (27.2)       & 59.45 (123.6)         & 99.59 (1.1)   & 97.4 (2.8)     & -                      & - \\
PNN \cite{rusu2016progressive}              & 75.25 (93.51)      & 58.96 (588)           & -             & -              & -                      & -\\
A-GEM \cite{chaudhry2018efficient}          & 54.38 (25.4+16)    & -                     & -             & -              & -                      \\
ER-RES \cite{chaudhry2019tiny}            & 66.78 (25.4+16)    & 57.32 (102.6+110.1)   & -             & -              & -                      & -\\
UCB \cite{ebrahimi2019uncertainty}          & -                  & -                     & 99.63 (2.2)   & 91.44 (2.2)    & 76.34 (32.8)           & -\\
ACL \cite{ACL_ECCV_2020}                    & 78.08 (25.1)       & 62.07 (113.81+8.5)    & 99.76 (1.6)   & 98.03 (2.4)    & 78.55 (16.5)           & -\\

Baseline                                    & 59.31 (52.06)     & 51.14 (112.26)        & 99.23 (1.28)  & 85.25 (0.67)   & 82.17 (12.76)          & 82.13 (15.65)\\

SAN (Ours)                                  & \textbf{71.73 (26.2)}      & \textbf{62.60 (108.27)}        & \textbf{99.65 (1.50)}  & \textbf{98.72 (2.2})    & \textbf{89.37 (16.24)}         & \textbf{83.31 (15.33)}\\
 \midrule  \midrule
Independent                                 & 69.53 (197.6)      & 63.02 (343.6)         & 99.78 (1.9)   & 98.91 (3.0)    & 89.60 (31.2)           & 93.22 (53.10)\\
\bottomrule
\end{tabular}
}
\end{center}
\caption{Accuracy (model size) comparison of different methods. Reply memory size is added with the model size where applicable. Accuracy and size are presented in percent and Megabyte (MB), respectively.}
\label{tab:acc}
\end{table*}

\section{Experiments}
\label{sec:experiments}


\subsection{Setup} 

\noindent\textbf{2D image datasets:} We conduct experiments on five different arrangements. We performed the experiments on 20-split CIFAR100 \cite{cifar100}, 20-split MiniImageNet \cite{miniimagenet}, 5-split MNIST, and 10 split Permuted MNIST. We also run the experiment on a more challenging dataset, a sequence of 5 datasets consisting of MNIST, CIFAR10, notMNIST, SVHN, and Fashion MNIST, to understand the performance of a complex task. Here we describe some statistics of those datasets:  (1) MNIST \cite{lecun-mnisthandwrittendigit-2010} and FashionMNIST \cite{xiao2017fashion}, PermutedMNIST \cite{lecun1998gradient} comprises 60000 training images and 10000 test images of $28\times28$ pixels grayscale images from 10 classes. (2) The MiniImageNet dataset contains 100 classes randomly chosen from the ImageNet \cite{deng2009imagenet} dataset, and each class includes 600 sample images of $84\times84$ color images. (3) NotMNIST dataset comprises $28\times28$ grayscale images. The dataset is divided into 18265 training samples and 459 test samples of alphabet `A' to `J', a total of 10 classes. (4) SVHN \cite{netzer2011reading} dataset contains 99289 real color images of size $32\times32$ pixels collected from Google Street View. Each image represents a number from $0-9$. The dataset is divided into 73257 train samples and 26032 test samples. (5) CIFAR10 is a 10 class dataset that contains 60000 images of different real-world objects. All images of the dataset are three channels and $32\times32$ pixels. 50000 data is allotted for the training set, and the rest 10000 images are used for the test set. Each class contains a total of 6000 images. (6) CIFAR100 is a similar dataset to CIFAR10, and it contains a total of 60000 color $32\times32$ pixel images distributed to 100 classes. Every class has 500 training and 100 test images. 

\noindent\textbf{Incremental settings:} In this paper, we follow the setting proposed by \cite{ACL_ECCV_2020}. In this setting, for MiniImageNet and CIFAR100, we split the datasets into 20 subsets. Each subset contains five classes and an equal number of training and validation samples. In the 5-split MNIST experiment, we divided the dataset into five tasks, and every task contains two classes. The Permuted MNIST dataset is partitioned into ten tasks. Finally, the sequence of 5 datasets comprises CIFAR10, MNIST, Fashion-MNIST, SVHN, and NotMNIST datasets. Each dataset contains 10 class labels. The distribution of the classes with tasks is randomly chosen during the run-time.

\noindent \textbf{Beyond 2D data:} We conduct 3D experiments with 40 classes of ModelNet40 \cite{modelnet2015} using PointNet \cite{pointcnn2018} as the 3D architecture. The dataset contains 9843 training and 2468 test 3D models. Based on the class names' ascending alphabetical order, we divide all 40 classes into four groups representing 
Class ID(1-10), Class ID(11-20), Class ID(21-30), and Class ID(31-40).

\noindent\textbf{Evaluation metrics:} For every incremental task, we measure the accuracy (in percent) to analyze the performance of each task. Finally, we report the mean accuracy.

\noindent\textbf{Validation strategy:} We randomly split the training set into two sets. Following \cite{ACL_ECCV_2020}, the first set contains 85\% of the training data to train the model, and the rest 15\% is used as validation data for all the datasets mentioned above.

\noindent \textbf{Implementation Details\footnote{Code and modes are available at: \url{TBA}}:} We used a small convolutional Neural Network for all our experiments except 5-split MNIST and 10-split Permuted MNIST. The backbone comprises three convolution layers, the adjustment network consists of 4 convolution layers, and the classifier is composed of a three-layer perceptron. However, the backbone consists of one convolution layer for 5-split MNIST and 10-split Permuted MNIST. The classification layer contains three fully connected layers with ReLU activation in between, and the adjustment network consists of only a single convolution layer. Adam is used to optimizing the network for all experiments, and we used categorical cross-entropy as the loss function.

\noindent\textbf{Comparison methods:} We compared our work with the following methods.
\textit{(1)} Published works: PNN \cite{rusu2016progressive}, HAT \cite{serra2018overcoming}, ER-RES \cite{chaudhry2019tiny}, A-GEM \cite{chaudhry2018efficient}, ACL \cite{ACL_ECCV_2020}, UCB \cite{ebrahimi2019uncertainty} etc.
\textit{(2)} Fine-Tune: This method fine-tunes the initial model without growing the network size in each incremental step. 
\textit{(3)} Baseline: Method described in Sec. \ref{sec:solution-overview}.
\textit{(4)} Independent: We trained a separate neural network for each task. The backbone and classifier networks are trained for each task. Therefore, this model provides the upper-bound performance.



\subsection{Overall Results}
\label{sec:rnd}



\noindent\textbf{With 2D image data:} In Table \ref{tab:acc}, we present accuracy (model size) comparison of different methods. The Fine-Tune method achieved inferior performance across datasets because the network size did not grow, aligning with newer task data to train the network. The baseline method beats Fine-Tune because it allows growing the network size by adding a newer classifier for new task data. Different published methods successfully outperform the common baseline in most cases by applying problem-specific strategies (like reply memory, specialized arch., regularization, etc.). Our proposed SAN model consistently beats the common baseline in accuracy and arch. size. It tells that SAN could serve as a stronger baseline than the current one while comparing with other works. The Independent method beats all methods because it allows training separate networks instead of a single network. Note, Independent is not any incremental learning solution. It is used to check the upper bound of performance.\\
\noindent\textbf{With 3D point cloud data:} There are no published results for 3D task incremental learning. Table \ref{tab:acc} also shows that the Independent model represents the average upper bound accuracy of 93.22\% with the arch. size of 53.10 MB for 4 splits of ModelNet40 \cite{modelnet2015} dataset. The Baseline model's size is 15.65 MB, with an average accuracy of 82.13\%. Finally, SAN outperforms the Baseline model attaining 83.31\% average accuracy with a smaller arch. of size 15.33 MB.



\subsection{Ablation studies}

\noindent\textbf{Impact of the order of incremental tasks:} Since our model trains the backbone and classifier only in the first task, the data used in Task 1 may impact overall performance. In Table \ref{tab:incremental_order}, we show the impact of the order of incremental tasks in both 2D and 3D domains.
Unlike CIFAR10, using the MNIST dataset as Task 1 training decreases the average accuracy for the Sequence of 5 datasets. This is because MNIST is a handwritten text dataset, whereas CIFAR10 is a dataset of natural images. Natural images seem to train the backbone better than handwritten text. On the other hand, we observe that the task order does not influence the 3D dataset because all tasks contain 3D models from the same dataset. The average accuracy does not decrease considerably when Class ID (11-20) or Class ID (21-30) are utilized for Task 1 training.

\begin{table}[!t]
\begin{minipage}{0.45\textwidth}
\centering \small
\scalebox{.75}{
\begin{tabular}{@{}cccccccc@{}}
\toprule
\textbf{2D datasets} & Task 1 & Task 2 & Task 3 & Task 4  & Task 5 & Average  \\ \midrule
 & CIFAR10 & MNIST & NotMNIST & SVHN  & F\_MNIST &   \\ 
CIFAR10    & 74.34  & 98.58  & 96.95  & 88.14  & 88.85  & 89.37 \\ 
\midrule
 & MNIST & CIFAR10 & NotMNIST & SVHN  & F\_MNIST &   \\ 
MNIST  & 62.96  & 99.36  & 90.31  & 91.50  & 90.31  & 86.88 \\ \bottomrule
\end{tabular}
}
\end{minipage}
\vspace*{0.1 cm}

\begin{minipage}{0.45\textwidth}
\centering \small
\scalebox{0.75}{
\begin{tabular}{@{}ccccccc@{}}
\toprule
\textbf{3D dataset} & Task 1 & Task 2 & Task 3 & Task 4  & Average  \\ \midrule
  & CID(11-20) & CID(1-10) & CID(21-30) & CID(31-40) & \\
CID(11-20) & 75.52 & 95.21  & 65.88 & 84.16 & 80.19 \\
\midrule
  & CID(21-30) & CID(1-10) & CID(11-20) & CID(31-40) & \\
CID(21-30)  & 67.82 & 97.01 & 73.13 & 83.29 & 80.31 \\ \bottomrule
\end{tabular}
}
\end{minipage}
\caption{Impact of incremental order. CID refers class ID.}
\label{tab:incremental_order}
\end{table}



\begin{figure}[!t]
    \centering
    \subfloat{{\includegraphics[scale=0.44]{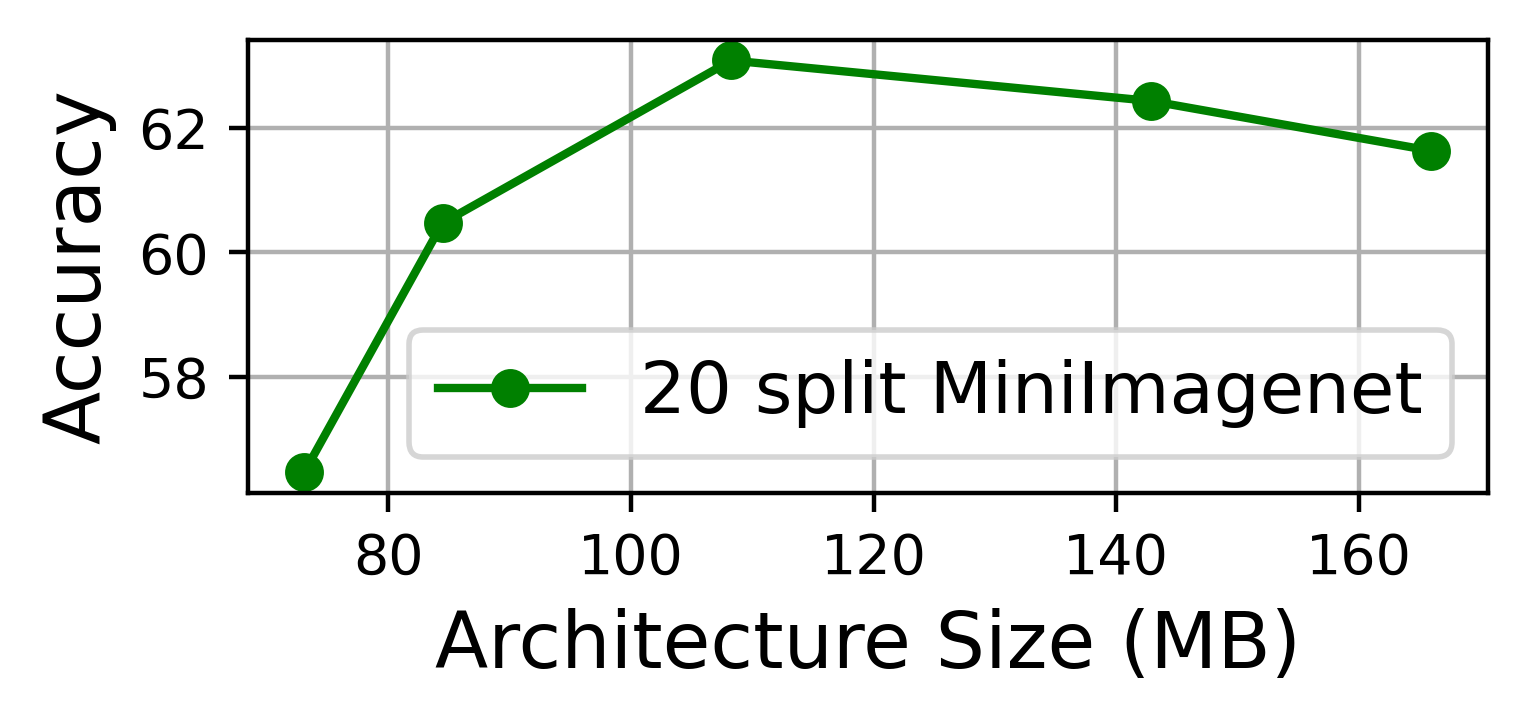} }}%
    \subfloat{{\includegraphics[scale=0.44]{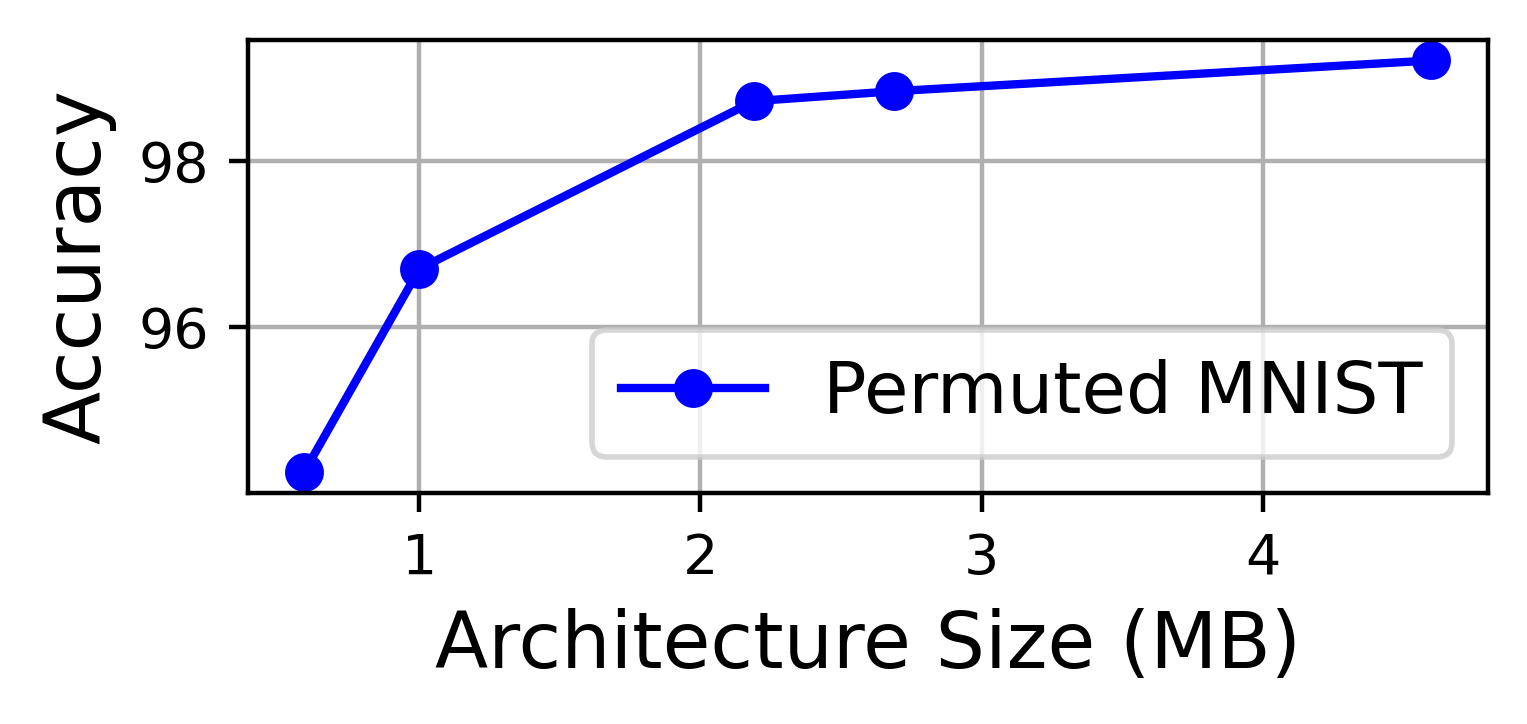} }}%
    \caption{Impact of varying architecture size.}%
    \label{fig:varying_size}%
\end{figure}


\noindent\textbf{Varying the model size:} In Fig. \ref{fig:varying_size}, we manipulate the adjustment network, $\mathcal{F}_t$ to vary the model size and record the impact of varying architecture size. To alter the model size, we bring variations on the kernel size of the convolutional layers. Hence, the Backbone, $B_1$ and the classifier, $\mathcal{C}_1$ remain unchanged. For 20 split MiniImagenet, starting with a smaller size of 73.07 MB, we experimented with larger models up to 166MB. The model achieves 56.47\% accuracy with the minor architecture, and the performance increases until it finds a sweet spot of 108.27 MB achieving 63.08\% accuracy. After this point, the model starts dropping performance. For permuted MNIST, we see a high slope until it reaches a size 2.2 MB with 98,72\% accuracy, and then the rise of performance becomes slow. This experiment tells that SAN can compete for state-of-the-art results by tweaking the model size.

\noindent\textbf{Visualization:} Fig. \ref{fig:tsne} shows the latent space of the common baseline and SAN method. For the baseline case (Fig. \ref{fig:tsne} (left)), after the first incremental step with Task 1, data \textcolor{orange} {orange} and \textcolor{green}{green} dot clusters are created. In the second incremental step, new clusters of instances are created (\textcolor{blue}{blue} and \textcolor{red}{red} dot) in the latent space. One can notice there are some inter-mixing of the same task instances (especially for Task 2) which creates confusion during test time. It happens because the backbone did not get a chance to update itself for new tasks. On the contrary, after training the first incremental step, SAN (Fig. \ref{fig:tsne} (right)) does not create any more clusters. It only adjusts newer task instances (\textcolor{blue}{blue} and \textcolor{red}{red} dot) to previously computed decision boundaries (\textcolor{orange} {orange} and \textcolor{green}{green} dot). One can notice there is almost no overlap between same task instances which eventually helps to improve the performance. This becomes possible because of conv. layers of adjustment network that creates better features helpful for the previously built classifiers.

\noindent\textbf{Use of memory instances:} SAN is not prone to forgetting issues. Even without using any memory instances, SAN can compete with memory-based task-incremental methods like \cite{buzzega2021rethinking, chaudhry2019tiny, guo2019improved, weston2014memory, pritzel2017neural, santoro2016one}.

\newcommand{\rulesep}{\unskip\ \vrule\ }
\begin{figure}[!t]
    \centering
    \subfloat{{\includegraphics[scale=0.17]{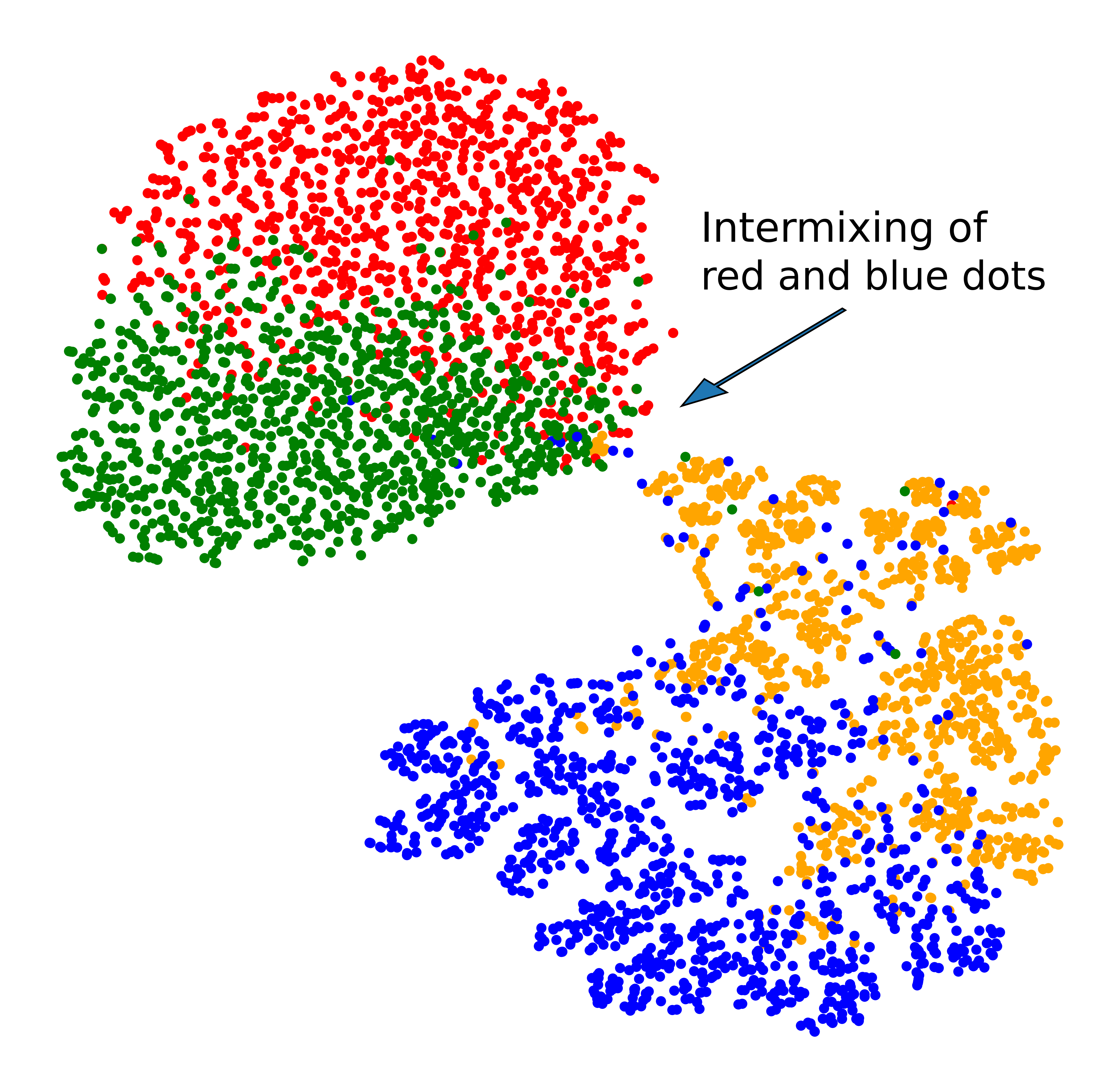} }}%
    \rulesep
    \subfloat{{\includegraphics[scale=0.17]{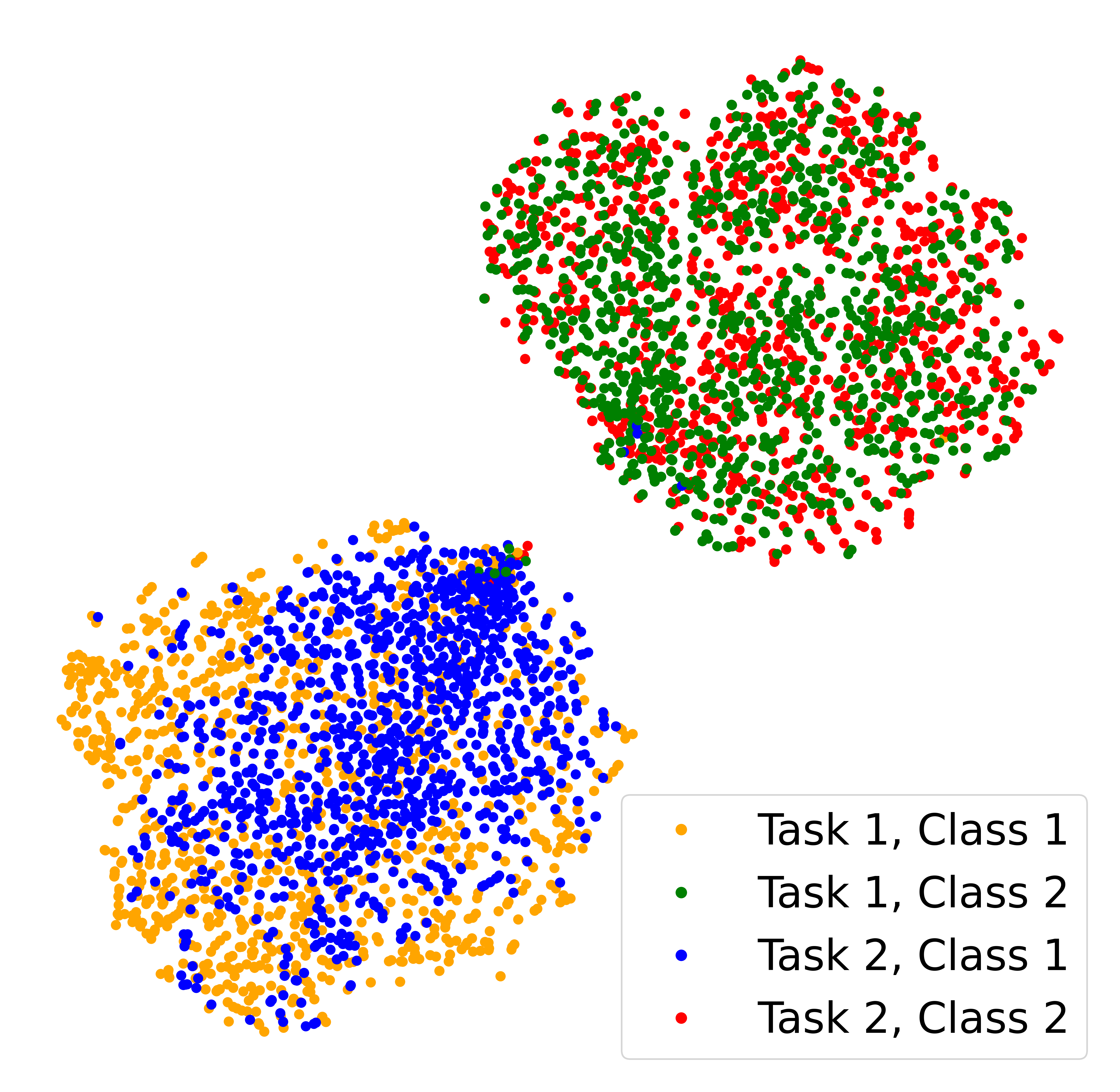} }}%
    \caption{2D tSNE visualization of the latent space on \textbf{(left)} common baseline and \textbf{(right)} our proposal, SAN.}%
    \label{fig:tsne}%
\end{figure}


\section{Conclusion}
This paper proposes a simple yet effective baseline method, SAN for task-incremental learning. Unlike existing methods, our proposal can maintain satisfactory performance using a small model size without any memory dependency. For new tasks, SAN adjusts a portion of its network to accommodate the latest batch of the data stream. By retaining the previously learned knowledge and only using its adjustable part, the network achieves near state-of-the-art performance. Finally, we evaluate our model performance on the benchmark dataset with state-of-the-art works. In addition, we also achieve consistent results on the 3D point cloud dataset. Our model exhibits extraordinary performance with the lowest architecture growth when learning from a continuously evolving data stream and task throughout life. Therefore, the takeaway message is that one should consider SAN as a strong baseline for a fair comparison among task-incremental learning approaches.

\bibliographystyle{IEEEtran}
\bibliography{references}

\begin{thebibliography}{10}
\providecommand{\url}[1]{#1}
\csname url@samestyle\endcsname
\providecommand{\newblock}{\relax}
\providecommand{\bibinfo}[2]{#2}
\providecommand{\BIBentrySTDinterwordspacing}{\spaceskip=0pt\relax}
\providecommand{\BIBentryALTinterwordstretchfactor}{4}
\providecommand{\BIBentryALTinterwordspacing}{\spaceskip=\fontdimen2\font plus
\BIBentryALTinterwordstretchfactor\fontdimen3\font minus
  \fontdimen4\font\relax}
\providecommand{\BIBforeignlanguage}[2]{{%
\expandafter\ifx\csname l@#1\endcsname\relax
\typeout{** WARNING: IEEEtran.bst: No hyphenation pattern has been}%
\typeout{** loaded for the language `#1'. Using the pattern for}%
\typeout{** the default language instead.}%
\else
\language=\csname l@#1\endcsname
\fi
#2}}
\providecommand{\BIBdecl}{\relax}
\BIBdecl

\bibitem{buzzega2021rethinking}
P.~Buzzega, M.~Boschini, A.~Porrello, and S.~Calderara, ``Rethinking experience
  replay: a bag of tricks for continual learning,'' in \emph{2020 25th
  International Conference on Pattern Recognition (ICPR)}.\hskip 1em plus 0.5em
  minus 0.4em\relax IEEE, 2021, pp. 2180--2187.

\bibitem{chaudhry2019tiny}
A.~Chaudhry, M.~Rohrbach, M.~Elhoseiny, T.~Ajanthan, P.~K. Dokania, P.~H. Torr,
  and M.~Ranzato, ``On tiny episodic memories in continual learning,''
  \emph{arXiv preprint arXiv:1902.10486}, 2019.

\bibitem{guo2019improved}
Y.~Guo, M.~Liu, T.~Yang, and T.~Rosing, ``Improved schemes for episodic
  memory-based lifelong learning,'' \emph{arXiv preprint arXiv:1909.11763},
  2019.

\bibitem{weston2014memory}
J.~Weston, S.~Chopra, and A.~Bordes, ``Memory networks,'' \emph{arXiv preprint
  arXiv:1410.3916}, 2014.

\bibitem{pritzel2017neural}
A.~Pritzel, B.~Uria, S.~Srinivasan, A.~P. Badia, O.~Vinyals, D.~Hassabis,
  D.~Wierstra, and C.~Blundell, ``Neural episodic control,'' in
  \emph{International Conference on Machine Learning}.\hskip 1em plus 0.5em
  minus 0.4em\relax PMLR, 2017, pp. 2827--2836.

\bibitem{santoro2016one}
A.~Santoro, S.~Bartunov, M.~Botvinick, D.~Wierstra, and T.~Lillicrap,
  ``One-shot learning with memory-augmented neural networks,'' \emph{arXiv
  preprint arXiv:1605.06065}, 2016.

\bibitem{aljundi2017expertgate}
R.~Aljundi, P.~Chakravarty, and T.~Tuytelaars, ``Expert gate: Lifelong learning
  with a network of experts,'' in \emph{Proceedings of the IEEE Conference on
  Computer Vision and Pattern Recognition}, 2017, pp. 3366--3375.

\bibitem{chaudhry2018riemannian}
A.~Chaudhry, P.~K. Dokania, T.~Ajanthan, and P.~H. Torr, ``Riemannian walk for
  incremental learning: Understanding forgetting and intransigence,'' in
  \emph{Proceedings of the European Conference on Computer Vision (ECCV)},
  2018, pp. 532--547.

\bibitem{ebrahimi2019uncertainty}
S.~Ebrahimi, M.~Elhoseiny, T.~Darrell, and M.~Rohrbach, ``Uncertainty-guided
  continual learning with bayesian neural networks,'' \emph{arXiv preprint
  arXiv:1906.02425}, 2019.

\bibitem{serra2018overcoming}
J.~Serra, D.~Suris, M.~Miron, and A.~Karatzoglou, ``Overcoming catastrophic
  forgetting with hard attention to the task,'' in \emph{International
  Conference on Machine Learning}.\hskip 1em plus 0.5em minus 0.4em\relax PMLR,
  2018, pp. 4548--4557.

\bibitem{rusu2016progressive}
A.~A. Rusu, N.~C. Rabinowitz, G.~Desjardins, H.~Soyer, J.~Kirkpatrick,
  K.~Kavukcuoglu, R.~Pascanu, and R.~Hadsell, ``Progressive neural networks,''
  \emph{arXiv preprint arXiv:1606.04671}, 2016.

\bibitem{chaudhry2018efficient}
A.~Chaudhry, M.~Ranzato, M.~Rohrbach, and M.~Elhoseiny, ``Efficient lifelong
  learning with a-gem,'' \emph{arXiv preprint arXiv:1812.00420}, 2018.

\bibitem{ACL_ECCV_2020}
S.~Ebrahimi, F.~Meier, R.~Calandra, T.~Darrell, and M.~Rohrbach, ``Adversarial
  continual learning,'' in \emph{Computer Vision--ECCV 2020: 16th European
  Conference, Glasgow, UK, August 23--28, 2020, Proceedings, Part XI 16}.\hskip
  1em plus 0.5em minus 0.4em\relax Springer, 2020, pp. 386--402.

\bibitem{modelnet2015}
Z.~Wu, S.~Song, A.~Khosla, F.~Yu, L.~Zhang, X.~Tang, and J.~Xiao, ``{3D
  ShapeNets: A deep representation for volumetric shapes},'' in
  \emph{Proceedings of the IEEE/CVF Conference on Computer Vision and Pattern
  Recognition (CVPR)}, 2015.

\bibitem{cifar100}
A.~Krizhevsky, G.~Hinton \emph{et~al.}, ``Learning multiple layers of features
  from tiny images,'' 2009.

\bibitem{miniimagenet}
O.~Vinyals, C.~Blundell, T.~Lillicrap, D.~Wierstra \emph{et~al.}, ``Matching
  networks for one shot learning,'' \emph{Advances in neural information
  processing systems}, vol.~29, pp. 3630--3638, 2016.

\bibitem{lecun-mnisthandwrittendigit-2010}
\BIBentryALTinterwordspacing
Y.~LeCun and C.~Cortes, ``{MNIST} handwritten digit database,'' 2010. [Online].
  Available: \url{http://yann.lecun.com/exdb/mnist/}
\BIBentrySTDinterwordspacing

\bibitem{lecun1998gradient}
Y.~LeCun, L.~Bottou, Y.~Bengio, and P.~Haffner, ``Gradient-based learning
  applied to document recognition,'' \emph{Proceedings of the IEEE}, vol.~86,
  no.~11, pp. 2278--2324, 1998.

\bibitem{netzer2011reading}
Y.~Netzer, T.~Wang, A.~Coates, A.~Bissacco, B.~Wu, and A.~Y. Ng, ``Reading
  digits in natural images with unsupervised feature learning,'' 2011.

\bibitem{xiao2017fashion}
H.~Xiao, K.~Rasul, and R.~Vollgraf, ``Fashion-mnist: a novel image dataset for
  benchmarking machine learning algorithms,'' \emph{arXiv preprint
  arXiv:1708.07747}, 2017.

\bibitem{10.1007/978-3-319-46493-0_37}
Z.~Li and D.~Hoiem, ``Learning without forgetting,'' in \emph{Computer Vision
  -- ECCV 2016}, B.~Leibe, J.~Matas, N.~Sebe, and M.~Welling, Eds.\hskip 1em
  plus 0.5em minus 0.4em\relax Cham: Springer International Publishing, 2016,
  pp. 614--629.

\bibitem{8100070}
\BIBentryALTinterwordspacing
S.~Rebuffi, A.~Kolesnikov, G.~Sperl, and C.~H. Lampert, ``icarl: Incremental
  classifier and representation learning,'' in \emph{2017 IEEE Conference on
  Computer Vision and Pattern Recognition (CVPR)}.\hskip 1em plus 0.5em minus
  0.4em\relax Los Alamitos, CA, USA: IEEE Computer Society, jul 2017, pp.
  5533--5542. [Online]. Available:
  \url{https://doi.ieeecomputersociety.org/10.1109/CVPR.2017.587}
\BIBentrySTDinterwordspacing

\bibitem{kemker2018fearnet}
\BIBentryALTinterwordspacing
R.~Kemker and C.~Kanan, ``Fearnet: Brain-inspired model for incremental
  learning,'' in \emph{International Conference on Learning Representations},
  2018. [Online]. Available: \url{https://openreview.net/forum?id=SJ1Xmf-Rb}
\BIBentrySTDinterwordspacing

\bibitem{v.2018variational}
\BIBentryALTinterwordspacing
C.~V. Nguyen, Y.~Li, T.~D. Bui, and R.~E. Turner, ``Variational continual
  learning,'' in \emph{International Conference on Learning Representations},
  2018. [Online]. Available: \url{https://openreview.net/forum?id=BkQqq0gRb}
\BIBentrySTDinterwordspacing

\bibitem{liu2020mnemonics}
Y.~Liu, Y.~Su, A.-A. Liu, B.~Schiele, and Q.~Sun, ``Mnemonics training:
  Multi-class incremental learning without forgetting,'' in \emph{Proceedings
  of the IEEE/CVF conference on Computer Vision and Pattern Recognition}, 2020,
  pp. 12\,245--12\,254.

\bibitem{lopez2017gradient}
D.~Lopez-Paz and M.~Ranzato, ``Gradient episodic memory for continual
  learning,'' \emph{Advances in neural information processing systems},
  vol.~30, pp. 6467--6476, 2017.

\bibitem{xiang2019incremental}
Y.~Xiang, Y.~Fu, P.~Ji, and H.~Huang, ``Incremental learning using conditional
  adversarial networks,'' in \emph{Proceedings of the IEEE/CVF International
  Conference on Computer Vision}, 2019, pp. 6619--6628.

\bibitem{lee2019generative}
S.~Lee and J.-G. Baek, ``Generative pseudorehearsal strategy for fault
  classification under an incremental learning,'' in \emph{2019 IEEE
  International Conference on Computational Science and Engineering (CSE) and
  IEEE International Conference on Embedded and Ubiquitous Computing
  (EUC)}.\hskip 1em plus 0.5em minus 0.4em\relax IEEE, 2019, pp. 138--140.

\bibitem{wang2017growing}
Y.-X. Wang, D.~Ramanan, and M.~Hebert, ``Growing a brain: Fine-tuning by
  increasing model capacity,'' in \emph{Proceedings of the IEEE Conference on
  Computer Vision and Pattern Recognition}, 2017, pp. 2471--2480.

\bibitem{lee2019overcoming}
K.~Lee, K.~Lee, J.~Shin, and H.~Lee, ``Overcoming catastrophic forgetting with
  unlabeled data in the wild,'' in \emph{Proceedings of the IEEE/CVF
  International Conference on Computer Vision}, 2019, pp. 312--321.

\bibitem{yoon2018lifelong}
J.~Yoon, J.~Lee, E.~Yang, and S.~J. Hwang, ``Lifelong learning with dynamically
  expandable network,'' in \emph{International Conference on Learning
  Representations}.\hskip 1em plus 0.5em minus 0.4em\relax International
  Conference on Learning Representations, 2018.

\bibitem{mallya2018piggyback}
A.~Mallya, D.~Davis, and S.~Lazebnik, ``Piggyback: Adapting a single network to
  multiple tasks by learning to mask weights,'' in \emph{Proceedings of the
  European Conference on Computer Vision (ECCV)}, 2018, pp. 67--82.

\bibitem{tang2020graph}
B.~Tang and D.~S. Matteson, ``Graph-based continual learning,'' \emph{arXiv
  preprint arXiv:2007.04813}, 2020.

\bibitem{riemer2018learning}
M.~Riemer, I.~Cases, R.~Ajemian, M.~Liu, I.~Rish, Y.~Tu, and G.~Tesauro,
  ``Learning to learn without forgetting by maximizing transfer and minimizing
  interference,'' \emph{arXiv preprint arXiv:1810.11910}, 2018.

\bibitem{kirkpatrick2017overcoming}
J.~Kirkpatrick, R.~Pascanu, N.~Rabinowitz, J.~Veness, G.~Desjardins, A.~A.
  Rusu, K.~Milan, J.~Quan, T.~Ramalho, A.~Grabska-Barwinska \emph{et~al.},
  ``Overcoming catastrophic forgetting in neural networks,'' \emph{Proceedings
  of the national academy of sciences}, vol. 114, no.~13, pp. 3521--3526, 2017.

\bibitem{zenke2017continual}
F.~Zenke, B.~Poole, and S.~Ganguli, ``Continual learning through synaptic
  intelligence,'' in \emph{International Conference on Machine Learning}.\hskip
  1em plus 0.5em minus 0.4em\relax PMLR, 2017, pp. 3987--3995.

\bibitem{fernando2017pathnet}
C.~Fernando, D.~Banarse, C.~Blundell, Y.~Zwols, D.~Ha, A.~A. Rusu, A.~Pritzel,
  and D.~Wierstra, ``Pathnet: Evolution channels gradient descent in super
  neural networks,'' \emph{arXiv preprint arXiv:1701.08734}, 2017.

\bibitem{aljundi2018memory}
R.~Aljundi, F.~Babiloni, M.~Elhoseiny, M.~Rohrbach, and T.~Tuytelaars, ``Memory
  aware synapses: Learning what (not) to forget,'' in \emph{Proceedings of the
  European Conference on Computer Vision (ECCV)}, 2018, pp. 139--154.

\bibitem{goodfellow2013empirical}
I.~J. Goodfellow, M.~Mirza, D.~Xiao, A.~Courville, and Y.~Bengio, ``An
  empirical investigation of catastrophic forgetting in gradient-based neural
  networks,'' \emph{arXiv preprint arXiv:1312.6211}, 2013.

\bibitem{liu2021l3doc}
Y.~Liu, Y.~Cong, G.~Sun, T.~Zhang, J.~Dong, and H.~Liu, ``L3doc: Lifelong 3d
  object classification,'' \emph{IEEE Transactions on Image Processing},
  vol.~30, pp. 7486--7498, 2021.

\bibitem{chowdhury2021learning}
T.~Chowdhury, M.~Jalisha, A.~Cheraghian, and S.~Rahman, ``Learning without
  forgetting for 3d point cloud objects,'' in \emph{International
  Work-Conference on Artificial Neural Networks}.\hskip 1em plus 0.5em minus
  0.4em\relax Springer, 2021, pp. 484--497.

\bibitem{pointcnn2018}
Y.~Li, R.~Bu, M.~Sun, W.~Wu, X.~Di, and B.~Chen, ``Pointcnn: Convolution on
  x-transformed points,'' in \emph{NeurIPS}, 2018.

\bibitem{deng2009imagenet}
J.~Deng, W.~Dong, R.~Socher, L.-J. Li, K.~Li, and L.~Fei-Fei, ``Imagenet: A
  large-scale hierarchical image database,'' in \emph{2009 IEEE conference on
  computer vision and pattern recognition}.\hskip 1em plus 0.5em minus
  0.4em\relax Ieee, 2009, pp. 248--255.

\end{thebibliography}

\end{document}